\documentclass[10pt,twocolumn,letterpaper]{article}

\usepackage{cvpr}              

\usepackage{booktabs} 
\usepackage{caption}  
\usepackage{pifont}

\usepackage{wrapfig}
\usepackage{amssymb}
\usepackage{subcaption}
\usepackage[table]{xcolor}
\usepackage{xcolor}   
\usepackage{multirow}
\usepackage{makecell} 
\usepackage{url}
\usepackage{booktabs}
\usepackage{graphicx}
\usepackage{xspace}
\usepackage{amsmath,amsfonts,bm}
\usepackage{enumitem}

\definecolor{cvprblue}{rgb}{0.21,0.49,0.74}
\definecolor{iccvblue}{rgb}{0.21,0.49,0.74}
\definecolor{mitblue}{rgb}{0.88,0.95,0.96}
\definecolor{gold}{rgb}{0.75,0.6,0.12}
\colorlet{shadecolor}{gray!40}
\definecolor{mydarkred}{rgb}{0.8,0.02,0.02}

\newcolumntype{g}{>{\columncolor{mitblue}}c}
\newcolumntype{f}{>{\columncolor{mitblue}}l}
\newcolumntype{h}{>{\columncolor{mitblue}}r}
\newcolumntype{i}{>{\columncolor{gray}}c}

\def\modelname{JetViT\xspace}
\def\method{Post-Training Attention Search\xspace}

\usepackage[most]{tcolorbox}

\definecolor{cvprblue}{rgb}{0.21,0.49,0.74}
\usepackage[pagebackref,breaklinks,colorlinks,allcolors=cvprblue]{hyperref}

\def\paperID{41642} 
\def\confName{CVPR}
\def\confYear{2026}

\title{
JetViT: Efficient High-Resolution Vision Transformer with \\ Post-Training Attention Search
}

\author{
Dongyun Zou$^{1,*}$\quad Zhuoyang Zhang$^{1}$\quad Junyu Chen$^{3}$\quad Wenkun He$^{1}$\quad Qinhe Peng$^{2}$\quad
Hanrong Ye$^{3}$\\
Yao Lu$^{4}$\quad Hongxu Yin$^{3}$\quad Yu Wang$^{3}$\quad
Song Han$^{1,3}$\quad Han Cai$^{3,*}$\\[0.3em]
$^{1}$MIT\quad $^{2}$University of Pennsylvania\quad $^{3}$NVIDIA\quad $^{4}$Physical Intelligence
}

\begin{document}
\twocolumn[{
\maketitle
\begin{center}
    \captionsetup{type=figure}
    \includegraphics[width=\textwidth]{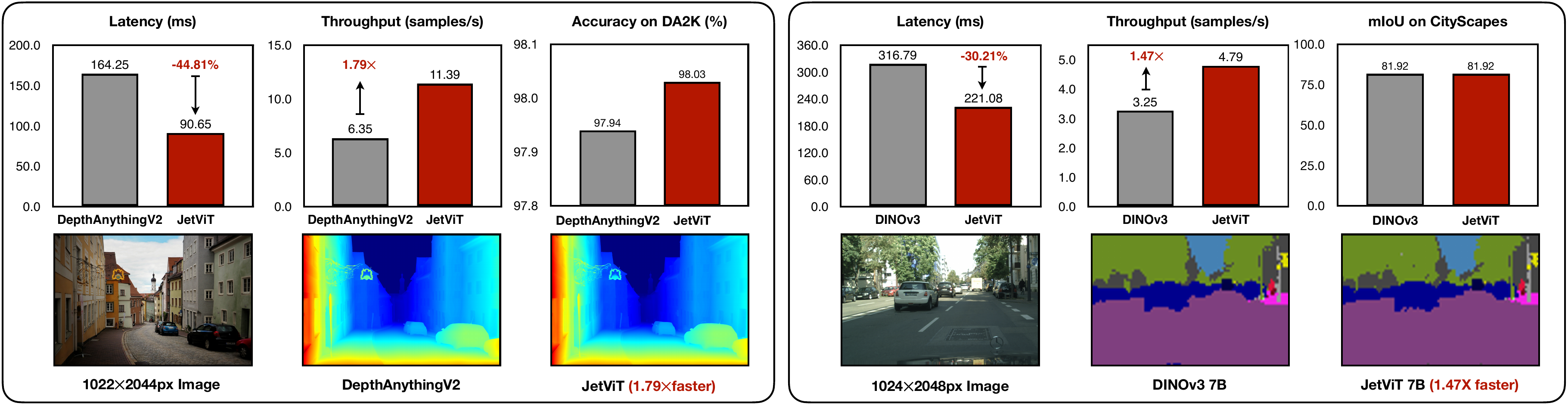}
    \captionof{figure}{\textbf{\modelname - Efficient Hybrid Attention Vision Transformers.} We transform state-of-the-art full-attention vision foundation models (e.g., DepthAnything, DINOv3) into efficient hybrid attention models using our cost-effective \emph{\method}. On DepthAnythingV2 models~\cite{depth_anything_v2}, \modelname achieves a $1.79\times$ increase in throughput and a 44.81\% reduction in latency without any loss in accuracy. On DINOv3 models~\cite{dinov3}, \modelname provides a $1.47\times$ throughput speedup and a 30.21\% latency reduction while maintaining comparable segmentation performance. All latency and throughput measurements are reported on an NVIDIA H100 GPU.}
\end{center}
}]
{
\renewcommand{\thefootnote}{}
\footnotetext{\hspace{-1.5em}$^*$Correspondence:\\ \texttt{dongyunzou03@gmail.com}\\ \texttt{hcai@nvidia.com}}
}
\begin{abstract}
We introduce \modelname, a novel family of hybrid-architecture Vision Transformer (ViT) models that match the accuracy of state-of-the-art full-attention vision foundation models while achieving substantially higher inference efficiency on high-resolution images. At the core of our approach is \method, a post-training acceleration framework that converts pre-trained full-attention ViTs into efficient hybrid-attention variants by identifying and replacing redundant full-attention blocks with linear or window-attention blocks. By inheriting the MLP and attention weights from the base model, \method efficiently explores the architectural design space through three key steps: (1) optimizing the linear-attention block design; (2) finding the best combination of linear-attention and window-attention blocks; and (3) identifying and preserving critical full-attention blocks. 
We evaluate \modelname on two representative high-resolution vision foundation models, DINOv3 and DepthAnythingV2. On the NVIDIA H100 GPU, \modelname achieves up to 1.79$\times$ higher throughput and up to 44.81\% lower latency without sacrificing accuracy. We will release our code and accelerated ViT models soon.
\end{abstract}
    
\section{Introduction}
\label{sec:intro}
\begin{figure*}[t]
    \centering
    \includegraphics[width=\linewidth]{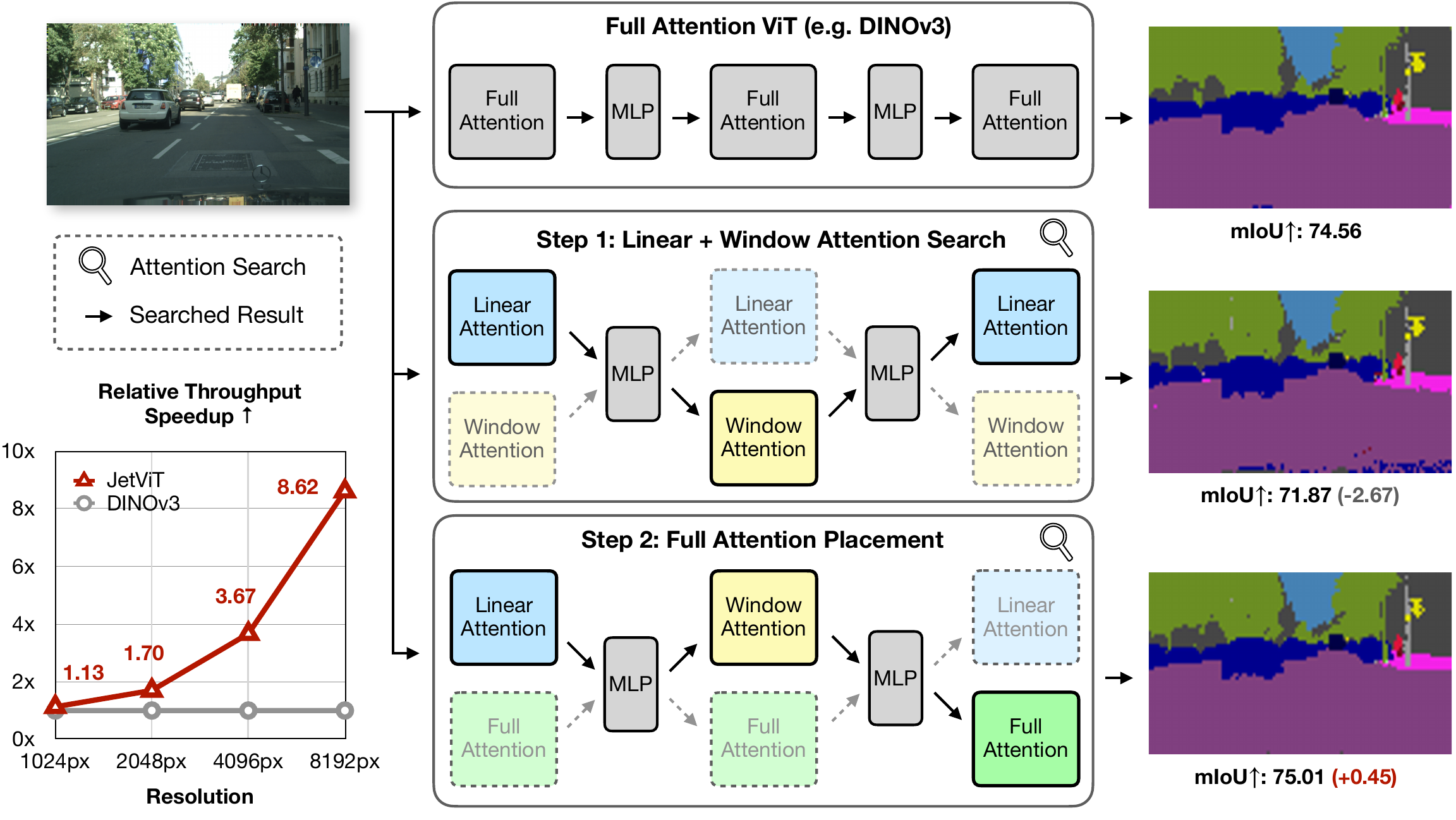}
    \caption{\textbf{\method Pipeline.} Our \method begins with a pre-trained full-attention Vision Transformer. We first search for the optimal combination of linear and window-attention blocks, producing an efficient ViT with $\text{O}(\text{N})$ complexity while retaining most of the performance of the original full-attention model. To close the gap in accuracy, we then perform a search to reintroduce a minimal number of full-attention blocks. The resulting hybrid ViT combines linear, window, and full-attention blocks, achieving accuracy comparable to the original model while delivering substantial speedups.
    }
    \label{fig:pipeline}
\end{figure*}

The Vision Transformer (ViT) \cite{vit} has rapidly emerged as a dominant paradigm in computer vision, achieving state-of-the-art performance across a wide range of applications. ViTs have proven to be highly effective as versatile backbones \cite{dinov3,mae,siglip2} and have delivered remarkable results in core vision tasks such as image classification \cite{wang2021pyramid, xia2022vision}, object detection \cite{detr, plain_detr}, and semantic segmentation \cite{sam, efficientvit_sam, cheng2021per}. Their capabilities further extend to specialized domains, including depth estimation \cite{depth_anything, depth_anything_v2} and generative image synthesis \cite{lit, sana}.

Despite their impressive performance, the core self-attention mechanism in standard Transformers \cite{attention, vit} incurs quadratic computational and memory complexity with respect to the input sequence length. This scaling poses significant challenges for processing high-resolution images and deploying models on resource-constrained devices. To address these limitations, extensive research has focused on developing more efficient Transformer variants. Key strategies include hierarchical architectures with window-based attention mechanisms \cite{swin} and various forms of linear-attention that reduce complexity to linear scaling \cite{linear_attention, efficientvit, flatten, meng2025polaformer}. These approaches aim to preserve the powerful global contextual modeling of Transformers while substantially improving efficiency.

However, the evaluation of these efficient architectures is often limited to small-scale benchmarks, such as ImageNet-1K classification \cite{imagenet}, and they are rarely tested on large-scale vision foundation models that require extensive pre-training on massive, sometimes proprietary, datasets \cite{dinov3}. 
This gap stems in part from the fact that pre-training a novel efficient architecture from scratch at such scale is computationally prohibitive, and access to high-quality web-scale private datasets is unavailable outside leading industrial AI labs. This raises a critical question: how can we close this gap and develop efficient models that match the accuracy of these leading vision foundation models while delivering significantly improved efficiency through better architectural design?

To address this challenge, we introduce \method, a post-training acceleration framework for ViTs. Rather than designing ViTs from scratch, \method starts with a pre-trained full-attention ViT and converts it into an efficient hybrid model containing a minimal number of full-attention blocks. The pipeline consists of three key steps: (i) Optimizing the linear-attention design: Inspired by hybrid LLM architectures, we combine a lightweight dynamic convolution with traditional ReLU-based linear attention, achieving better performance while introducing less overhead compared with previous methods. (ii) Finding the best combination of linear-attention and window-attention blocks. (iii) Identifying and preserving critical full-attention blocks.

Using \method, we derive \modelname, a new family of hybrid-architecture ViTs created by applying it to pre-trained full-attention vision foundation models. The accelerated JetViT models preserve the accuracy of the base model while retaining only two full-attention blocks, delivering a substantial efficiency boost on high-resolution images. For example, JetViT-DepthAnythingV2 achieves a 1.79$\times$ throughput improvement on an NVIDIA H100 GPU while maintaining comparable depth estimation accuracy to DepthAnythingV2. Similarly, JetViT-DINOv3 reduces latency by 30.21\% compared with DINOv3, while achieving the same segmentation accuracy on the Cityscapes dataset. We summarize our contributions as follows:
\begin{itemize}
    \item We propose the \modelname Linear Attention Block, a novel linear-attention design that combines ReLU-based linear attention with lightweight squeeze dynamic convolution. Compared with previous linear-attention blocks, it introduces minimal overhead while delivering improved performance.
    \item We propose \method, a two-stage distillation and search pipeline that efficiently converts a pre-trained full-attention ViT into a hybrid ViT composed of linear-attention, window-attention, and a small number of critical full-attention blocks.
    \item We validate our approach by applying \method on two representative high-resolution vision foundation models, DINOv3 and DepthAnythingV2. Our \modelname achieves up to 1.79$\times$ higher throughput and up to 44.81\% lower latency on an NVIDIA H100 GPU, without sacrificing accuracy.
\end{itemize}

\section{Related Work}
\label{sec:related}
\paragraph{High-Resolution Vision Foundation Models.} Recent advances in vision foundation models have greatly improved the ability to process high-resolution images, enabling more detailed and accurate visual understanding. The DINO series of models \cite{dino,dinov2,dinov3} represents a key milestone in this domain, with DINOv3 demonstrating exceptional performance in extracting rich, high-dimensional features from high-resolution inputs. Remarkably, DINOv3’s versatile feature representations require only a simple linear head to achieve strong results across multiple dense prediction tasks, including semantic segmentation and depth estimation, establishing an efficient paradigm for transfer learning. Building on this foundation, the DepthAnything series \cite{depth_anything,depth_anything_v2} adopts DINOv2 as its backbone and integrates a Dense Prediction Transformer (DPT) \cite{dpt} head to perform per-pixel depth prediction. This design effectively leverages the high-resolution feature extraction capabilities of vision foundation models, setting new benchmarks for monocular depth estimation in complex visual scenes. However, the high training cost of these foundation models prohibits the  exploration of efficient structures, which is the challenge we want to solve.

\paragraph{Efficient Vision Transformer Design.} Vision Transformers (ViTs) have demonstrated strong capabilities in extracting image features and have been successfully applied to numerous downstream tasks. However, the quadratic computational complexity of the self-attention mechanism limits their ability to directly process high-resolution images. To address this challenge, various approaches have been proposed. One line of work focuses on designing efficient attention mechanisms, such as linear-attention blocks \cite{linear_attention,efficientvit,sana,flatten,meng2025polaformer,vision_mamba} or other variants with computational complexity between $O(N)$ and $O(N^2)$ \cite{han2024agent,li2025radial}. Another common strategy is to restrict self-attention to local windows \cite{swin}. These approaches primarily target the pre-training stage. In contrast, our work focuses on the post-training stage \cite{he2025dc,chen2025dc}, which is significantly more cost-effective. Recent efforts have also attempted to convert full-attention ViTs into linear-attention variants \cite{lit,liu2024linfusion}. However, the architectures in these methods are often highly constrained and, as a result, cannot fully leverage the knowledge from the original full-attention teacher model.

\paragraph{Neural Architecture Search.} Neural Architecture Search (NAS)~\cite{once_for_all,cai2018proxylessnas,zoph2016neural,cai2018efficient} is a powerful approach for identifying model structures with optimal performance. Many efforts have focused on designing improved ViT architectures \cite{ni2021nasformer,chen2021glit,su2022vitas}. These methods typically operate at a very fine-grained search granularity, leading to enormous search spaces and high computational costs. In contrast, our work targets the search of attention blocks, dramatically simplifying the search process while enabling effective knowledge transfer from pre-trained full-attention models through weight inheritance.

\section{Method}
\label{sec:method}
\subsection{Motivation}
Training Vision Transformers (ViTs) from scratch incurs prohibitive computational costs, and state-of-the-art full-attention ViTs are often pretrained on massive proprietary datasets that are inaccessible to most researchers. This limitation severely constrains the exploration of novel efficient ViT architectures, with many proposed designs validated only on small-scale benchmarks (e.g., ImageNet-1K classification) rather than the diverse, high-resolution downstream tasks where full-attention ViTs excel. To bridge this gap, we propose \method, a post-training architecture search framework that efficiently converts pretrained full-attention ViTs into optimized hybrid models. Our method inherits all weights from the original ViT, eliminating the need for private datasets or costly re-pretraining. By leveraging public datasets and a two-stage distillation-and-search process (\cref{fig:pipeline}), \method produces highly efficient models that match the performance of their full-attention counterparts.

\subsection{\method}
\subsubsection{Linear-Attention Block Design}
\begin{table*}[t]
  \centering
  \setlength{\tabcolsep}{3pt}
  \begin{tabular}{lcccc}
    \toprule
    \multirow{2}{*}{Linear Attention Type} & 
    \makecell{1024$\times$2048px} & 
    Inference Latency & Inference Throughput& 
    \makecell{Train Step Time} \\
    & mIoU$\uparrow$ & milliseconds & samples/s & seconds \\
    \midrule
    Baseline: ReLU Linear Attention & 65.17 & 5.33 & 212.03 & 0.198 \\
    \midrule
    \hspace{1em}+ DWC on V  & 71.82  & \textbf{5.72} & \textbf{204.49} & \textbf{0.227} \\
    \hspace{1em}+ DWC on V + Focusing Factor\cite{flatten} & 72.48 & 6.00 & 191.09 & 0.312 \\
    \hspace{1em}+ Token-wise Dynamic DWC & - & 8.94 & 120.77 & - \\
    \hspace{1em}+ \textbf{Squeeze Dynamic DWC on V (Ours)} & \textbf{73.12} & \underline{5.88} & \underline{197.57} & \underline{0.232} \\
    \bottomrule
  \end{tabular}
  \caption{\textbf{Performance and Efficiency Comparison of Linear-Attention ViTs on the Cityscapes Dataset.} Using the full-attention ViT (DINOv3) as the teacher, we perform distillation training on a pure linear-attention ViT. Inference latency is measured on $1024 \times 2048$ px images, while training step time is measured using a batch of 1024 images at $256 \times 256$ px on a server with 8 H100 GPUs. Our linear-attention design introduces the least overhead compared with static depth-wise convolution (DWC) while achieving the best performance.}
  \label{tab:ablation_on_linear_attention}
\end{table*}

\begin{figure}[t]
    \centering
    \includegraphics[width=\linewidth]{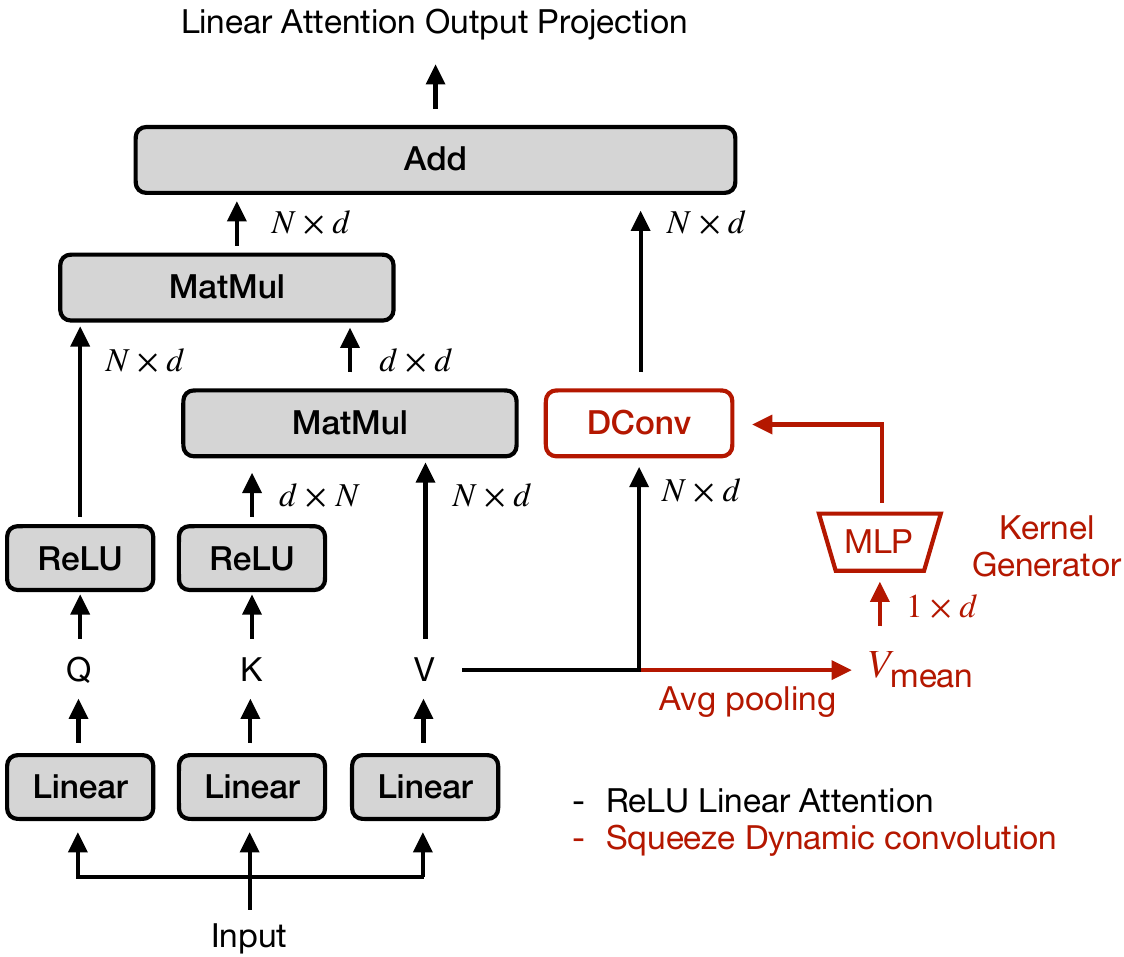}
    \caption{\textbf{JetViT Linear-Attention Block With Squeeze Dynamic Convolution.}}
    \label{fig:linear_attention}
\end{figure}

\paragraph{Preliminary.} Traditional full-attention introduces $\mathcal{O}(N^2)$ complexity when computing pairwise similarity:
\begin{equation}
\text{Sim}(Q,K) = \exp\left(\frac{QK^\top}{\sqrt{d}}\right).
\end{equation}

\noindent To address this issue, linear attention was introduced. It employs a kernel function $\phi(\cdot)$ to decompose the similarity computation as:
\begin{equation}
\text{Sim}(Q,K) = \phi(Q)\phi(K)^\top.
\end{equation}

\noindent By reordering the computations, linear attention avoids explicitly constructing the $N \times N$ matrix, thereby reducing the complexity to $\mathcal{O}(N)$.

However, using simple kernel functions such as ReLU often leads to a significant drop in model performance. To mitigate this, previous works have proposed two strategies to enhance linear attention \cite{flatten,meng2025polaformer}:
\begin{enumerate}
    \item Applying depth-wise convolution (DWC) on the $V$ matrix to capture local information.
    \item Replacing ReLU with carefully designed kernel functions, e.g., 
    \begin{equation}
    \phi(x) = \frac{\|x\|}{\|x\|^p} x^p,
    \end{equation}
    where $p$ is a focusing factor, to produce a sharper attention map.
\end{enumerate}

\noindent With these improvements, linear attention can be formulated as:
\begin{equation}
    O=\phi(Q)\phi(K)^TV+\text{DWC}(V)
\end{equation}
where DWC denotes depth-wise convolution.

\paragraph{JetViT Linear Attention Block.}
Our ablation studies on enhancements to simple ReLU linear attention reveal that while depth-wise convolution provides a substantial improvement, additional kernel functions offer negligible gains at a significant computational cost (e.g., 37\% slower training speed).

To improve linear-attention efficiency while minimizing overhead, we draw inspiration from dynamic convolution mechanisms recently employed in LLM architectures \cite{jet_nomotron,cai2024condition,wu2019pay}. These approaches typically use an MLP as a kernel generator to produce a 1D convolution kernel for each token. However, our initial attempt to directly replace the depth-wise convolution with such a token-wise dynamic convolution kernel resulted in prohibitive overhead (e.g., a 40\% drop in throughput).

Prior work on dynamic convolution in image processing \cite{wu2019pay} often generates a single dynamic kernel based on global features, which is then shared across all tokens. Inspired by this global-feature-based approach, we propose replacing the depth-wise convolution with a lightweight dynamic convolution, referred to as Squeeze Dynamic Convolution (see \cref{fig:linear_attention}). Specifically, we substitute the static weights of the depth-wise convolution with dynamically generated weights derived from the average pooling of the value matrix $V$. A compact two-layer MLP with SiLU \cite{silu} activation functions serves as the kernel generator.

As shown in \cref{tab:ablation_on_linear_attention}, our proposed linear-attention block outperforms traditional counterparts. By retaining the ReLU kernel function, it introduces negligible computational overhead compared to the original static depth-wise convolution variant. It is worth noting that while recent state-space models such as Vision Mamba \cite{vision_mamba} also provide an efficient alternative for modeling long-range dependencies, their core mechanism requires a bidirectional scan—a forward and backward pass—within each layer to aggregate global information, effectively doubling computational cost compared to our single-pass approach. Moreover, our linear-attention block is designed for direct weight initialization from pre-trained full-attention Vision Transformers, a step shown to be critical for accelerating convergence (\cref{sec:method_details}). Vision Mamba’s distinct state-space parameterization prevents this direct weight transfer, limiting its ability to fully leverage the benefits of inheriting pre-trained model knowledge.

\subsubsection{Linear and Window-Attention Search}
Alongside linear attention (LA), window attention (WA) is another widely used efficient attention mechanism. WA works by partitioning the feature map into non-overlapping windows and computing self-attention independently within each window. Typically, WA is alternated with full attention (FA), with FA responsible for capturing global context. This motivates our investigation into whether LA can effectively replace the computationally expensive FA for global information aggregation, enabling a hybrid architecture composed solely of LA and WA blocks.

\begin{table}[t]
    \centering
    \begin{tabular}{l|cc}
    \toprule
     Attention Types & mIoU & pAcc  \\
     \midrule
     Pure Full Attention & \textbf{68.74} & \textbf{93.81} \\
     \midrule
     Pure Linear Attention & 61.48 & 92.67 \\
     Pure Window Attention & 65.14 & 93.19 \\
     \midrule
     \makecell[l]{LA + WA Hybrid with \\ \method} & \underline{66.43} & \underline{93.40} \\
     \bottomrule
    \end{tabular}
    \caption{\textbf{Comparison of Linear-Probing Segmentation Results on Cityscapes ($512 \times 512$ px).} With \method, the optimal combination of linear and window attention can recover most of the full-attention model’s performance while outperforming ViTs that use only linear or only window attention.}
    \label{tab:attention_types}
\end{table}

We apply \method as illustrated in Step 1 of \cref{fig:pipeline}. Starting from a pretrained full-attention ViT (DINOv3 base), we first construct a supernetwork containing two attention blocks per layer. Using the full-attention ViT as the teacher, we perform distillation on the supernetwork. After training, we employ beam search \cite{beam_search} to identify the optimal combination of linear-attention and window-attention blocks. The results are presented in \cref{tab:attention_types}. With careful selection, a ViT composed of linear and window-attention blocks can recover most of the performance of the full-attention ViT while surpassing models that use only linear or only window-attention blocks.

\subsubsection{Full-Attention Placement}
However, a performance gap between the searched efficient ViT and the teacher model still remains. Previous studies on hybrid language models \cite{jet_nomotron} indicate that preserving a few full-attention blocks is crucial for improving model accuracy. To close this gap, we continue to leverage our efficient \method to identify the optimal placement of full-attention blocks.

As illustrated in Step 2 of \cref{fig:pipeline}, we construct a supernetwork containing two attention blocks per layer: an efficient attention block (searched in Step 1) and a full-attention block. Using the full-attention ViT as the teacher, we perform distillation training and employ beam search to determine the best placement of the full-attention blocks. Remarkably, we find that only two full-attention blocks are sufficient to recover the model’s performance. This finding is further validated by extended experiments presented in \cref{sec:exp}.

\begin{figure}[t]
    \centering
    \includegraphics[width=1.0\linewidth]{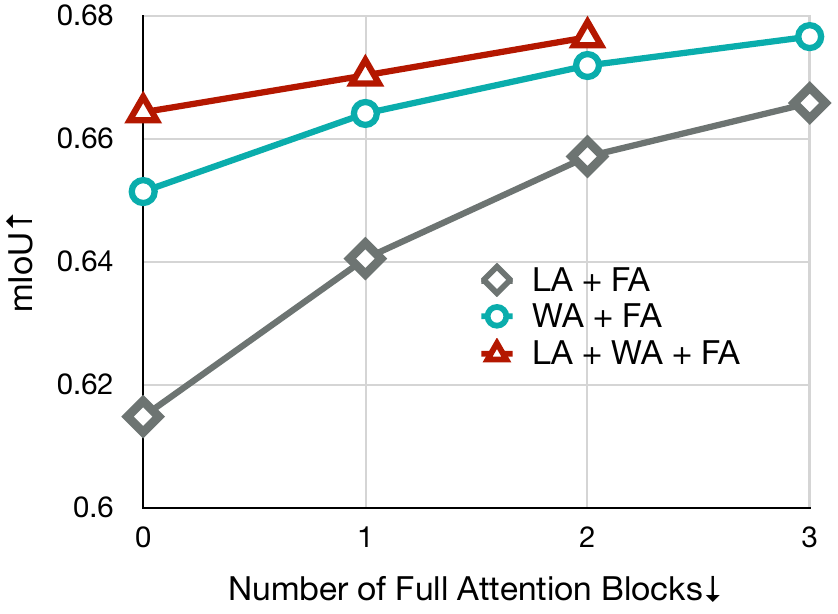}
    \caption{\textbf{Comparison of Segmentation Results on Cityscapes ($512 \times 512$ px).} With the same full-attention budget, combining linear, window, and full-attention blocks achieves the highest mIoU. This highlights the importance of first identifying an efficient attention block combination before determining the placement of full-attention blocks.}
    \label{fig:ablation_on_attention_types}
    \vspace{-10pt}
\end{figure}

We also examine whether Step 1 is necessary by conducting an ablation on linear- and window-attention search. Specifically, we directly apply \method using two attention types (e.g., linear-attention + full-attention or window-attention + full-attention) and compare the results with those from the two-stage \method. The results are shown in \cref{fig:ablation_on_attention_types}. We find that combining all three attention types further reduces the number of full-attention blocks needed to recover the hybrid model’s performance, yielding a more efficient model structure.

\subsubsection{Implementation Details}
\label{sec:method_details}
\begin{figure}[t]
    \centering
    \includegraphics[width=1.0\linewidth]{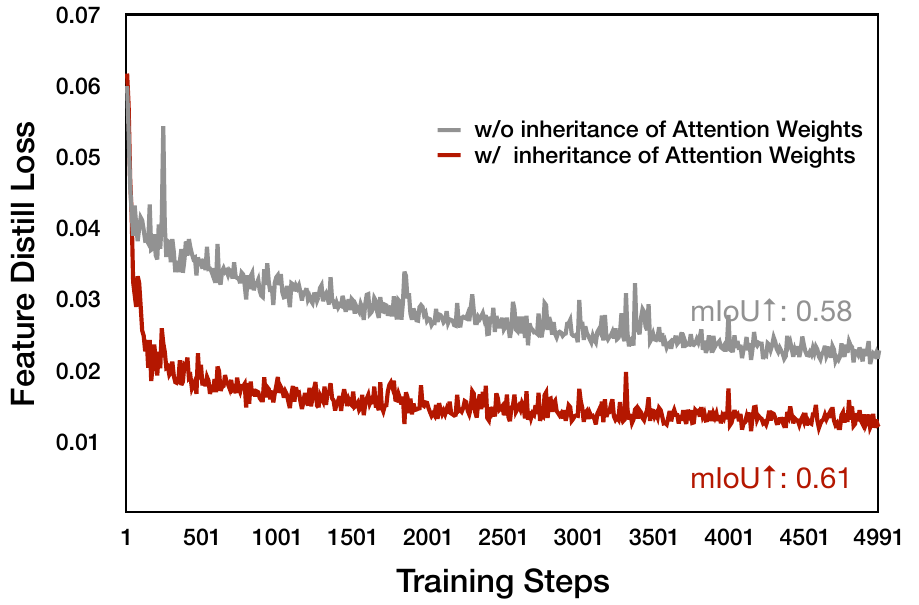}
    \vspace{-15pt}
    \caption{\textbf{Training Curve Comparison: Effect of Inheriting Full-Attention Weights.} The gray curve inherits only MLP weights, while the red curve inherits all weights from the base model.}
    \vspace{-15pt}
    \label{fig:ablation_on_inherit_weight}
\end{figure}

\paragraph{Inherit Weights from Base Model.} Previous studies on converting full-attention Transformers into linear or hybrid models often inherit only the MLP weights, while keeping the $Q$, $K$, $V$, and $O$ matrices in the efficient attention blocks randomly initialized \cite{jet_nomotron,liu2024linfusion}. We show that inheriting all attention weights ($\text{W}_\text{Q}, \text{W}_\text{K}, \text{W}_\text{V}, \text{W}_\text{O}$) is critical for maintaining performance. Specifically, we train a pure linear-attention ViT using a full-attention ViT as the teacher. The loss curves in \cref{fig:ablation_on_inherit_weight} demonstrate that inheriting the full-attention weights (red curve) achieves a +3\% mIoU on Cityscapes segmentation with a linear head, compared to inheriting only the MLP weights (gray curve).

\vspace{-10pt}
\paragraph{Distillation Training.} Following \cite{once_for_all}, we train a supernetwork and perform a search to efficiently identify the optimal attention combination. At each training step, a subnetwork is randomly selected for forward and backward passes. Feature distillation is applied to the features used for the downstream task (e.g., only the last layer when using a linear head for segmentation, or four intermediate layers when using a DPT head to predict a depth map). 

\vspace{-10pt}
\paragraph{Beam Search.} We perform beam search using the downstream task’s metric (e.g., mIoU, DA2K) as the direct search objective. The search proceeds in a greedy, stage-wise manner, guided by the known efficiency hierarchy of attention types (linear-attention $>$ window-attention $>$ full-attention). In Step 1, we start with an all-linear-attention configuration and incrementally replace linear-attention blocks with window-attention until performance gains plateau. In Step 2, we take the best architecture from Step 1 and progressively substitute its efficient attention blocks with full-attention blocks, stopping once the model’s performance converges to that of the original full-attention teacher.
\begin{table*}[t]
    \centering
    \begin{tabular}{lcccccccc}
    \toprule
    \multirow{2}{*}{Model}& \multirow{2}{*}{Size(Spec)} & Latency & Throughput & \multicolumn{2}{c}{Cityscapes 1024x2048px} & \multicolumn{2}{c}{ADE20K 512x512px} \\
    \cmidrule(lr){3-3} \cmidrule(lr){4-4} \cmidrule(lr){5-6} \cmidrule(lr){7-8}
    && ms $\downarrow$ & samples/s $\uparrow$ & mIoU$\uparrow$ & pAcc$\uparrow$ & mIoU$\uparrow$ & pAcc$\uparrow$ \\
    \midrule
    AM-RADIO v3\cite{heinrich2025radiov25improvedbaselinesagglomerative} & Base & 12.08 & 87.59 & \textbf{76.43} & \textbf{95.37} & 49.10 & 81.87 \\
    DINOv3\cite{dinov3} & Base & 12.86 & 81.52 & 74.56 & 95.19 & \textbf{50.32} & \textbf{82.59} \\
    JetViT-DINOv3 & Base (0 FA) & 7.37 & 153.94 & 71.87 & 94.75 & 48.84 & 82.16 \\
    JetViT-DINOv3 & Base (1 FA) & 7.69 & 145.45 & 74.50 & 95.20 & 49.36 & 82.40 \\
    JetViT-DINOv3 & Base (2 FA) & 8.27 & 133.68 & \underline{75.01} & \underline{95.30} & \underline{49.85} & \underline{82.53} \\
    \midrule
    AM-RADIO v3\cite{heinrich2025radiov25improvedbaselinesagglomerative} & Large & 33.24 & 31.15 & 78.50 & 95.78 & 51.20 & 82.61 \\
    DINOv3\cite{dinov3} & Large & 35.07 & 29.53 & \underline{79.84} & \underline{96.07} & \textbf{52.57} & \underline{82.74} \\
    JetViT-DINOv3 & Large (2 FA) & 18.77 & 57.90 & \textbf{79.88} & \textbf{96.08} & \underline{52.31} & \textbf{82.75}\\
    \midrule
    DINOv3\cite{dinov3} & 7B & 316.79 & 3.26 & \textbf{81.92} & \underline{96.44} & \underline{54.72} & \underline{83.55}\\
    JetViT-DINOv3 & 7B (2 FA) & 221.08 & 4.80 & \textbf{81.92} & \textbf{96.45} & \textbf{54.86} & \textbf{83.70} \\
    \bottomrule
    \end{tabular}
    \caption{\textbf{Semantic Segmentation Results on Cityscapes and ADE20K.} (\textbf{Best results}, \underline{Second-best results}) \modelname-DINOv3 achieves performance comparable to state-of-the-art vision backbone models (DINOv3 \cite{dinov3} and AM-RADIO \cite{heinrich2025radiov25improvedbaselinesagglomerative}) while providing a significant speedup.}
    \label{tab:segmentation}
    \vspace{-10pt}
\end{table*}

\section{Experiments}
\label{sec:exp}
\subsection{Setup}
\paragraph{Datasets.} To ensure diversity in our training data, we used a mixture of widely used open-source datasets, including SA1B \cite{sam}, ImageNet21K \cite{imagenet21k}, BDD100K \cite{bdd100k}, Google Landmarks \cite{google_landmarks}, Places365 \cite{places365}, Pexels \cite{pexels}, Cityscapes \cite{cityscapes}, and COCO \cite{coco}. We did not balance the dataset ratios but simply concatenated them.

\paragraph{Speed Test Setting.} We measure latency and throughput on a server equipped with eight NVIDIA H100 80GB HBM3 GPUs, two Intel Xeon Platinum 8462Y+ CPUs, and 2 TB of RAM. Our environment is based on PyTorch 2.8. For full-attention blocks, we use PyTorch’s native scaled\_dot\_product\_attention. The linear-attention blocks are implemented entirely in PyTorch, without any custom kernel optimizations.

\subsection{Semantic Segmentation}
We validate \modelname’s ability to extract image features without training on specific downstream tasks. Starting from DINOv3’s base, large, and 7B models, we convert them into efficient hybrid models (referred to as \modelname-DINOv3) using \method. Following the DINOv3 protocol, we train a simple linear head on the frozen backbone for semantic segmentation. Linear probing results provide an indicator of the quality of the extracted features, as shown in \cref{tab:segmentation}.

Performance is measured using mean Intersection over Union (mIoU) and pixel Accuracy (pAcc) on the CityScapes \cite{cityscapes} and ADE20K \cite{ade20k} datasets. We compare \modelname-DINOv3 with state-of-the-art vision backbones, including DINOv3 \cite{dinov3} and AM-RADIO \cite{heinrich2025radiov25improvedbaselinesagglomerative}. The linear head is trained for 5,000 steps using cross-entropy loss. Optimization is performed with the Adam optimizer, a learning rate of $1\times10^{-4}$, $\beta$ values of (0.9, 0.999), and a batch size of 1024.

The results demonstrate that \modelname-DINOv3 can extract visual features of comparable quality to the original DINOv3 models on both high- and low-resolution images.

\vspace{-2pt}
\subsection{Monocular Depth Estimation}
We further evaluate \modelname’s capabilities on a more complex task: monocular depth estimation. We start from the state-of-the-art DepthAnythingV2 model \cite{depth_anything_v2}, which uses a full-attention Vision Transformer as the backbone and a Dense Prediction Transformer (DPT) \cite{dpt} to predict pixel-wise inverse relative depth maps. DepthAnythingV2 fully leverages the ViT backbone features by extracting four intermediate layers and feeding them into the DPT head.

\begin{table*}[t]
    \centering
    \setlength{\tabcolsep}{2pt}
    \begin{tabular}{lcccccccccc}
    \toprule
    \multirow{2}{*}{Model} &\multirow{2}{*}{Size(Spec)} & Latency  & Throughput & DA2K & \multicolumn{2}{c}{DIODE} & \multicolumn{2}{c}{Sintel} & \multicolumn{2}{c}{CityScapes} \\
    \cmidrule(lr){3-3} \cmidrule(lr){4-4} \cmidrule(lr){5-5}  \cmidrule(lr){6-7} \cmidrule(lr){8-9} \cmidrule(lr){10-11}
    & & ms $\downarrow$ & samples/s $\uparrow$ & Acc$\uparrow$ & $\delta_1$$\uparrow$ & Abs Rel $\downarrow$ & $\delta_1$$\uparrow$ & Abs Rel $\downarrow$ & $\delta_1$$\uparrow$ & Abs Rel $\downarrow$ \\
    \midrule
    MiDaS V3 DPT\cite{midas} & Large(Swin) & 39.28 & 27.14 & 75.04 & 0.696 & 0.262 & 0.598 & 0.342 & 0.743 & 0.186 \\
    DepthAnythingV2\cite{depth_anything_v2} & Large & 60.76 & 14.20 & 97.60 & \textbf{0.749} & \underline{0.231} & \textbf{0.730} & \textbf{0.222} & \underline{0.872} & 0.111 \\
    JetViT-DepthAnything & Large(0 FA) & 29.70 & 34.97 & 97.74 & 0.743 & \textbf{0.228} & 0.722 & 0.229 & \underline{0.872} & 0.110 \\
    JetViT-DepthAnything & Large(1 FA) & 31.04 & 33.72 & \underline{97.76} & \underline{0.744} & 0.229 & \underline{0.726} & 0.227 & 0.871 & \underline{0.109} \\
    JetViT-DepthAnything & Large(2 FA) & 32.63 & 32.13 & \textbf{97.84} & \textbf{0.749} & \textbf{0.228} & \textbf{0.730} & \underline{0.226} & \textbf{0.876} & \textbf{0.108} \\
    \midrule
    DepthAnythingV2\cite{depth_anything_v2} & Giant & 164.25 & 6.35 & \underline{97.94} & \textbf{0.751} & \underline{0.234} & \underline{0.737} & \textbf{0.210} & \underline{0.876} & \underline{0.111} \\
    JetViT-DepthAnything & Giant(2 FA) & 90.65 & 11.39 & \textbf{98.03} & \underline{0.750} & \textbf{0.231} & \textbf{0.741} & \textbf{0.210} & \textbf{0.879} & \textbf{0.109} \\
    \bottomrule
    \end{tabular}
    \caption{\textbf{Zero-Shot Depth Estimation Results.} (\textbf{Best results}, \underline{Second-best results}) DepthAnything employs a full-attention Transformer as its backbone, while MiDaS V3 uses a Swin Transformer \cite{swin}, which is more efficient but suffers a notable drop in performance. In contrast, \modelname matches the performance of the state-of-the-art DepthAnythingV2 on these high-resolution datasets, while delivering a substantial speedup.}
    \label{tab:depth_estimation}
    \vspace{-10pt}
\end{table*}
\begin{figure*}[t]
    \centering
    \includegraphics[width=1.0\linewidth]{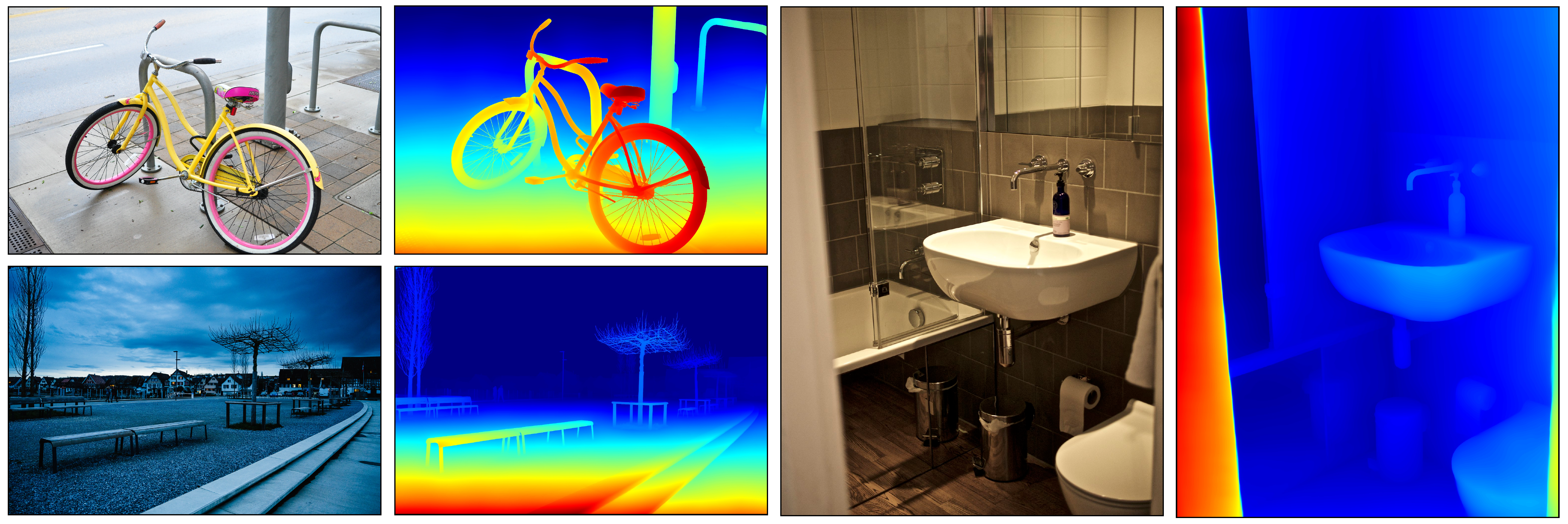}
    \caption{\textbf{Qualitative Results of \modelname-DepthAnythingV2.} Our model successfully inherits two key strengths from DepthAnythingV2: (i) high precision in capturing fine-grained details, as evidenced by the clearly reconstructed bicycle spokes and distant tree branches in the left scene; and (ii) robustness in handling complex surfaces, such as transparent glass and reflective mirrors in the right scene.}
    \label{fig:depth_visualization}
\end{figure*}

After performing the two-stage \method, we further fine-tune our model following the same training strategy as DepthAnythingV2. We use the largest DepthAnythingV2 model, which was purely trained on synthetic datasets, to generate pseudo depth labels for unlabeled real-image datasets, including SAM, BDD100K, Google Landmarks, Places365, Pexels, and COCO, and train our model solely on these pseudo labels.

The results are shown in \cref{tab:depth_estimation}. We evaluate using standard metrics, $\delta_1$ and absolute relative error (Abs Rel), on three high-resolution datasets: DIODE \cite{diode}, Sintel \cite{sintel}, and CityScapes \cite{cityscapes}. Images are used at their original high resolutions without resizing. Full metric definitions are provided in \cite{depth_anything}. Additionally, we measure the DA2K metric, proposed by DepthAnythingV2, which evaluates performance in complex environments (e.g., reflective or transparent objects). For DA2K images, we resize the short edge to 1022 pixels while preserving the aspect ratio.

Compared with state-of-the-art relative depth estimation models, DepthAnythingV2 \cite{depth_anything_v2} and MiDaS \cite{midas}, \modelname-DepthAnythingV2 achieves comparable or superior performance on high-resolution datasets. While MiDaS uses a Swin Transformer backbone and is more efficient than DepthAnythingV2, our \modelname-DepthAnythingV2 employs fewer global full-attention layers than MiDaS, surpassing it in both performance and efficiency.

Qualitative results are presented in \cref{fig:depth_visualization}. Visualizations of \modelname-DepthAnythingV2 Giant’s predictions demonstrate that, through \method, the model successfully inherits DepthAnythingV2’s strengths: (i) capturing fine-grained details, and (ii) maintaining robustness in complex environments.

\subsection{Analysis on Searched Model Structure}
\begin{figure*}[t!]
    \vspace{-10pt}
    \centering
    \includegraphics[width=1.0\linewidth]{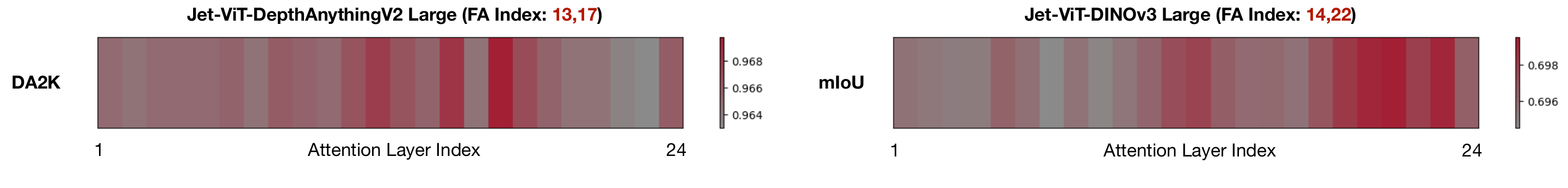}
    \caption{\textbf{Full-Attention Placement Search Results on a 24-Layer ViT Large.} Each grid cell represents the search objective value for replacing the corresponding layer with full-attention. Higher values indicate a greater gain from re-inserting full-attention.
    }
    \label{fig:heat_map}
    \vspace{-10pt}
\end{figure*}

Our analysis of \method’s results shows that the optimal placement of critical full-attention layers depends on how the Vision backbone’s features are utilized. For instance, DepthAnythingV2 extracts features from four intermediate ViT layers and feeds them into a DPT head for depth prediction, whereas DINOv3, for semantic segmentation, directly uses the final backbone layer to train a linear head. As illustrated in \cref{fig:heat_map}, the middle-to-deeper layers are most important for \modelname-DepthAnythingV2, with full-attention placed at layers 13 and 17. In contrast, for \modelname-DINOv3, the final layers are more critical, leading the search to select layers 14 and 22. These observations offer valuable guidance for designing future hybrid ViT architectures.

A further consistent pattern is that the first attention block is always linear-attention, suggesting that linear-attention may function analogously to convolutional layers in traditional CNNs, extracting fundamental image features in the shallow layers.

\subsection{Additional Results}
We provide additional experiments in the supplementary material, including: (i) a computational cost analysis showing that \method requires only $\sim$1/68 of the pre-training cost for the DINOv3-7B model; (ii) generalization results on object detection (COCO) and image/video classification (ImageNet, UCF101); and (iii) efficiency benchmarks on edge devices (NVIDIA RTX 3090 and Jetson GB10), where \modelname achieves up to 2.4$\times$ speedup at 2K resolution.
\section{Conclusion}
\label{sec:conclusion}
We introduce \modelname, a new family of hybrid-architecture Vision Transformers that matches the performance of full-attention models while delivering significantly improved efficiency, particularly for high-resolution images. \modelname is enabled by two key innovations: (1) \method, an efficient strategy for converting full-attention ViTs into hybrid models through weight inheritance, distillation and beam search. This approach streamlines the exploration of efficient ViT architectures by facilitating the rapid integration of state-of-the-art efficient attention blocks with well-pretrained full-attention ViTs. (2) A newly designed linear-attention block that combines lightweight dynamic convolution with traditional linear-attention, contributing to faster inference in our hybrid models and reducing the number of full-attention blocks needed to recover model performance.

Extensive experiments demonstrate that \modelname achieves substantial speedups without sacrificing accuracy. Moreover, \method significantly reduce the cost of efficient architecture exploration. We believe this framework will accelerate innovation by enabling rapid application of newly designed efficient ViT blocks.

{
    \small
    \bibliographystyle{ieeenat_fullname}
    \bibliography{main}
}
\end{document}